\documentclass[10pt,twocolumn,letterpaper]{article}

 \usepackage{cvpr}              

\usepackage[T1]{fontenc}

\definecolor{cvprblue}{rgb}{0.21,0.49,0.74}
\usepackage[pagebackref,breaklinks,colorlinks,allcolors=cvprblue]{hyperref}
\usepackage{tcolorbox}
\usepackage{graphicx}
\usepackage{multirow} 
\def\paperID{*****} 
\def\confName{CVPR}
\def\confYear{2026}

\title{SpatialScout: Can VLM Agents Actively Explore and Map 3D Scenes?}

\author{
Zhaoxi Zhang$^{1}$\thanks{Project Lead.} \thanks{Equal Contribution}  \quad
Hao Liang$^{1}$\footnotemark[2] \quad
Linxuan Wu$^{2}$\footnotemark[2] \quad
Yiming Xu$^{1}$\footnotemark[2] \quad
Yifu Yuan$^{2}$\footnotemark[2] \quad
Kun Liang$^{1}$ \\
Chenming Tang$^{1}$ \quad
Meiyi Qiang$^{1}$ \quad
Wentao Zhang$^{1}$ \quad
Weikang Li$^{1}$\thanks{Corresponding author.} \quad
Yunfang Wu$^{1}$\footnotemark[3] \\
$^{1}$\,Peking University \qquad
$^{2}$\,Tianjin University \\
\href{mailto:zhaoxizhang25@stu.pku.edu.cn}{zhaoxizhang25@stu.pku.edu.cn}
}

\begin{document}
\maketitle

\begin{abstract}
    Spatial intelligence requires agents to actively acquire visual evidence and integrate observations into a coherent representation of the three-dimensional world. However, spatial question answering provides limited evidence of whether models can construct accurate metric maps through autonomous exploration. We introduce \textbf{SpatialScout}, a benchmark for evaluating vision-language model (VLM) agents through active object-level 3D mapping. SpatialScout comprises 80 diverse indoor scenes containing 3,140 target objects. Given an object inventory, RGB observations, and camera calibration and poses, agents control a camera to explore each scene and estimate object centers in a shared coordinate frame. This task requires sustained coordination of viewpoint selection, cross-view association, geometric estimation, and map revision under bounded interaction budgets. Our evaluation jointly measures localization accuracy and target coverage, using object-scale overlap scores that penalize missing predictions. Experiments with frontier VLMs reveal substantial limitations: the strongest evaluated model, Claude Opus 5.5, achieves a mean overlap of 41.71\%, with only 47.23\% of targets exceeding an IoU threshold of 0.5 and 3.25\% exceeding 0.9. These results expose a substantial gap between recovering coarse object layouts and constructing complete, precise metric maps. SpatialScout provides a quantitative testbed for developing agents that build reliable 3D spatial representations through active perception. Code is available at \textit{https://github.com/JesseZZZZZ/SpatialScout}.
\end{abstract}

\begin{figure}[!t]
    \centering
    \includegraphics[width=1\linewidth]{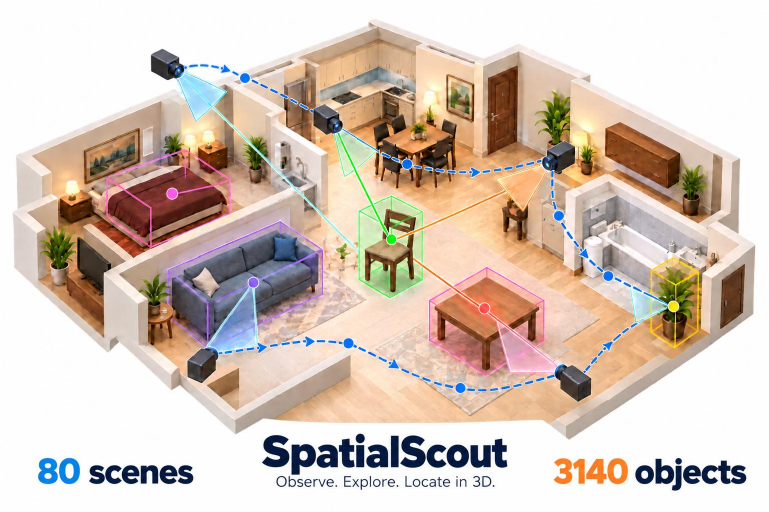}
    \caption{Overview of SpatialScout. A VLM agent observes, explores, and localizes the position of objects in a 3D environment.}
    \label{fig:teasor}
\end{figure}
\section{Introduction}
\label{sec:introduction}

\begin{table*}[t]
    \centering
    \caption{Comparison with spatial intelligence and scene understanding benchmarks. T/F: true/false; MC: multiple choice; NA: numerical answer; OE: open-ended.}
    \label{tab:benchmark_comparison}
    \footnotesize
    \setlength{\tabcolsep}{2pt}
    \renewcommand{\arraystretch}{1.18}
    \hfuzz=0pt
    \begin{tabular*}{\textwidth}{@{\extracolsep{\fill}}p{0.15\textwidth}p{0.18\textwidth}p{0.22\textwidth}cccc@{}}
        \toprule
        \multirow{2}{0.15\textwidth}{\textbf{Benchmark}} &
        \multirow{2}{0.18\textwidth}{\textbf{Output format}} &
        \multirow{2}{0.22\textwidth}{\textbf{Primary evaluation target}} &
        \textbf{Active} &
        \textbf{Multi-round} &
        \textbf{Explicit spatial} &
        \textbf{Direct 3D} \\
        & & &
        \textbf{interaction} &
        \textbf{env. feedback} &
        \textbf{construction} &
        \textbf{coordinate scoring} \\
        \midrule
        VSR~\cite{vsr} & T/F & Answer correctness & $\times$ & $\times$ & $\times$ & $\times$ \\
        BLINK~\cite{blink} & MC & Answer correctness & $\times$ & $\times$ & $\times$ & $\times$ \\
        VSI-Bench~\cite{vsi} & MC, NA & Answer correctness & $\times$ & $\times$ & $\times$ & $\times$ \\
        OpenEQA~\cite{openeqa} & OE & Answer correctness & $\times$ & $\times$ & $\times$ & $\times$ \\
        E3VS-Bench~\cite{e3vs} & OE & Answer correctness & $\checkmark$ & $\checkmark$ & $\times$ & $\times$ \\
        ESI-Bench~\cite{esi} & T/F, MC, NA & Answer correctness & $\checkmark$ & $\checkmark$ & $\times$ & $\times$ \\
        S$^{3}$-Bench~\cite{s3} & MC, NA & Answer correctness & $\times$ & $\times$ & $\times$ & $\times$ \\
        Theory of Space~\cite{theoryofspace} & OE & Map quality & $\checkmark$ & $\checkmark$ & $\checkmark$ & $\times$ \\
        IR3D-Bench~\cite{ir3d} & 3D Scene Parameters & Layout and spatial relations & $\times$ & $\times$ & $\checkmark$ & $\times$ \\
        \midrule
        \noalign{\begingroup
            \color{cvprblue!12}\hrule height 12pt\vskip-12pt
        \endgroup}
        \textbf{SpatialScout} & \textbf{3D Coordinates} & \textbf{Localization \& Coverage} & $\boldsymbol{\checkmark}$ & $\boldsymbol{\checkmark}$ & $\boldsymbol{\checkmark}$ & $\boldsymbol{\checkmark}$ \\
        \bottomrule
    \end{tabular*}
\end{table*}

Spatial intelligence is fundamental to embodied agents' ability to understand and act in 3D environments~\cite{openeqa,conceptgraphs}.
It requires agents to actively observe their surroundings to resolve occlusion and perspective ambiguities, infer object locations and spatial relationships, and integrate evidence across views into a consistent 3D representation~\cite{hartley2004,esi}.
Evaluating this entire process is central to assessing the spatial capabilities of vision-language models (VLMs).
Existing spatial intelligence benchmarks widely adopt image- and video-based question answering (QA) to assess counting, direction, distance, and spatial relations~\cite{vsi,site}.
Although these evaluations have revealed substantial model limitations, answer accuracy imposes only limited constraints on spatial understanding.
Finite answer choices allow models to arrive at correct answers through guessing or elimination; language priors and biases in answer distributions may also enable correct responses based on non-spatial cues~\cite{goyal2017vqa,agrawal2018priors}.
More importantly, spatial QA does not fully capture real-world demands: agents must autonomously gather evidence, update spatial representations, and use them to guide subsequent actions~\cite{openeqa,theoryofspace}.
Evaluating frontier VLMs therefore requires end-to-end assessment of this multi-step process, beyond answer accuracy alone.

Bridging this gap requires evaluating autonomous information acquisition and spatial representation construction within a single task, forming a closed loop of observation, reasoning, and decision-making.
Recent benchmarks such as ESI-Bench~\cite{esi} and S$^{3}$-Bench~\cite{s3} introduce active interaction into spatial QA, while Theory of Space~\cite{theoryofspace} examines the construction of discrete cognitive maps.
Our key advance is to make the construction of object maps in continuous 3D space through active interaction a direct evaluation objective for spatial intelligence.
Models must autonomously acquire visual evidence and recover the 3D positions of a target set in a common coordinate frame.
This task couples active perception, cross-view association, geometric estimation, and scene modeling: the current spatial estimate determines what to observe next, while the selected viewpoint shapes the evidence available thereafter.
Through spatial construction beyond QA, feedback-driven active exploration, and end-to-end task evaluation, we examine whether VLMs can transform local observations into accurate and complete metric spatial representations.

To this end, we introduce \textbf{SpatialScout}, a benchmark for evaluating the spatial intelligence of vision-language agents through active object-level 3D mapping.
Starting from an initial observation and a target inventory, an agent autonomously explores the scene and progressively constructs a 3D map of target object centers in a common coordinate frame.
The benchmark has three core features:
\textcircled{\scriptsize 1}~\textbf{Explicit spatial construction.}
Models must output a set of 3D object positions, enabling quantitative evaluation of spatial representation quality through localization accuracy and target coverage.
\textcircled{\scriptsize 2}~\textbf{Closed-loop active exploration.}
Models select subsequent viewpoints based on existing observations and the current map, then update position estimates using new evidence, forming a feedback loop between the active acquisition of perceptual evidence and spatial understanding.
\textcircled{\scriptsize 3}~\textbf{Long-horizon end-to-end capability evaluation.}
Models coordinate observation planning, object recognition, cross-view association, geometric estimation, and map updates across multiple interaction steps, with overall performance measured by final map quality.
This sustained coordination goes beyond QA tasks with short reasoning chains and better reflects real-world spatial understanding.

SpatialScout comprises 80 indoor scenes and 3,140 target objects, with 30--40 targets per scene, spanning diverse spatial layouts, object scales, and repeated category instances.
The benchmark provides a unified set of camera-control and geometric assistance tools, along with camera intrinsics and poses, and requires models to infer object centers from autonomously acquired visual evidence.
Evaluation jointly measures geometric estimation error, localization accuracy relative to object scale, and target coverage.
We also record map updates under a limited interaction budget to analyze how object-level spatial representations emerge as exploration proceeds.

We evaluate frontier VLMs, including GPT-6 Astra and Claude Opus~5.5.
The results show that models can recover parts of object layouts through autonomous exploration, but comprehensive coverage and precise localization remain challenging.
Models differ substantially in target omissions and center localization accuracy, while the proportion of targets meeting stringent scale-relative accuracy criteria remains low.
These findings indicate that recognizing objects or making coarse spatial judgments does not necessarily translate into constructing reliable metric maps.
By exposing these limitations through measurable localization and coverage metrics, SpatialScout provides a unified benchmark for studying how models construct and understand the 3D world through autonomous interaction.

\section{SpatialScout Construction}
\label{sec:method}

SpatialScout evaluates active object-center mapping by VLM agents.
Given an object inventory and calibrated camera poses, an agent selects RGB views, associates objects across observations, and maintains a three-dimensional map of their centers.
The benchmark combines a fixed scene catalog, a shared interaction interface, and object-level evaluation to measure the accuracy and completeness of the resulting map (Fig.~\ref{fig:main}).

\subsection{Task Definition}
\label{sec:method_task}
\begin{table}[t]
\centering
\small
\setlength{\tabcolsep}{4pt}
\caption{Shared agent interface. No tool reveals ground-truth target positions or evaluation scores.}
\begin{tabular}{@{}p{0.49\columnwidth}p{0.47\columnwidth}@{}}
\toprule
\textbf{Tool} & \textbf{Effect or output} \\
\midrule
\texttt{list\_entities} & List targets and metadata. \\
\texttt{get\_camera\_pose} & Return the camera pose. \\
\texttt{get\_camera\_intrinsics} & Return camera calibration. \\
\texttt{set\_camera\_position} & Set the camera position. \\
\texttt{move\_camera} & Translate the camera. \\
\texttt{rotate\_camera} & Rotate the camera. \\
\texttt{capture\_image} & Capture an RGB frame. \\
\texttt{triangulate\_\allowbreak observations} & Estimate a point from multiple views. \\
\texttt{submit\_positions} & Update the center map. \\
\texttt{finish\_mapping} & End the mapping episode. \\
\bottomrule
\end{tabular}

\label{tab:method_tools}
\end{table}
\paragraph{Inventory and map.}
A scene contains $N$ targets with public inventory $\mathcal{E}=\{(i,c_i,g_i)\}_{i=1}^{N}$: instance identifier $i$, category $c_i$, and equivalence group $g_i$.
During an episode, the agent maintains estimates $\widehat{\mathcal{P}}=\{\widehat{\mathbf{p}}_i:i\in S\}$ for submitted identifiers $S\subseteq\{1,\ldots,N\}$.
Evaluation compares multisets of center positions within each group, with identifiers serving as submission slots for individual estimates.

\paragraph{Geometric target.}
Each target center is the midpoint of the tight world-axis-aligned bounding box (AABB) of its rendered geometry.
For an articulated object, the box encloses the union of its rendered link geometries.
Let $\mathbf{b}_i^{-},\mathbf{b}_i^{+}$ denote the box corners and let $\mathbf{o}_0,R_0$ be the initial camera's world position and camera-to-world rotation.
The reference center in meters is
\begin{equation}
    \mathbf{p}_i=R_0^{\top}\left(\frac{\mathbf{b}_i^{-}+\mathbf{b}_i^{+}}{2}-\mathbf{o}_0\right).
    \label{eq:method_center}
\end{equation}
The box is constructed in world coordinates; Eq.~\eqref{eq:method_center} then expresses its midpoint in the task frame.
This definition gives each asset a reproducible localization target, including assets whose internal origins are offset from their visible geometry.
Center inference combines correspondence geometry with the object's observed spatial extent: a triangulated surface feature and the AABB center are generally different points.

\paragraph{Coordinate frame and inputs.}
The task uses a fixed right-handed frame with origin at the initial camera, $+x$ along its initial optical forward direction, $+y$ to its initial left, and $+z$ along its initial up direction.
All absolute map positions and returned camera poses use this frame.
An initially pitched camera gives a task up axis tilted relative to world vertical.
The agent receives RGB images, the inventory, camera intrinsics, poses, and the target-region radius $r=\max_i\|\mathbf{p}_i\|_2$.
Reference centers, boxes, and asset geometry remain in the evaluator.
The supplied camera geometry and scene-level search bound support metric estimation, while the agent selects views, establishes correspondences, and infers object centers.
\begin{figure}
    \centering
    \includegraphics[width=1\linewidth]{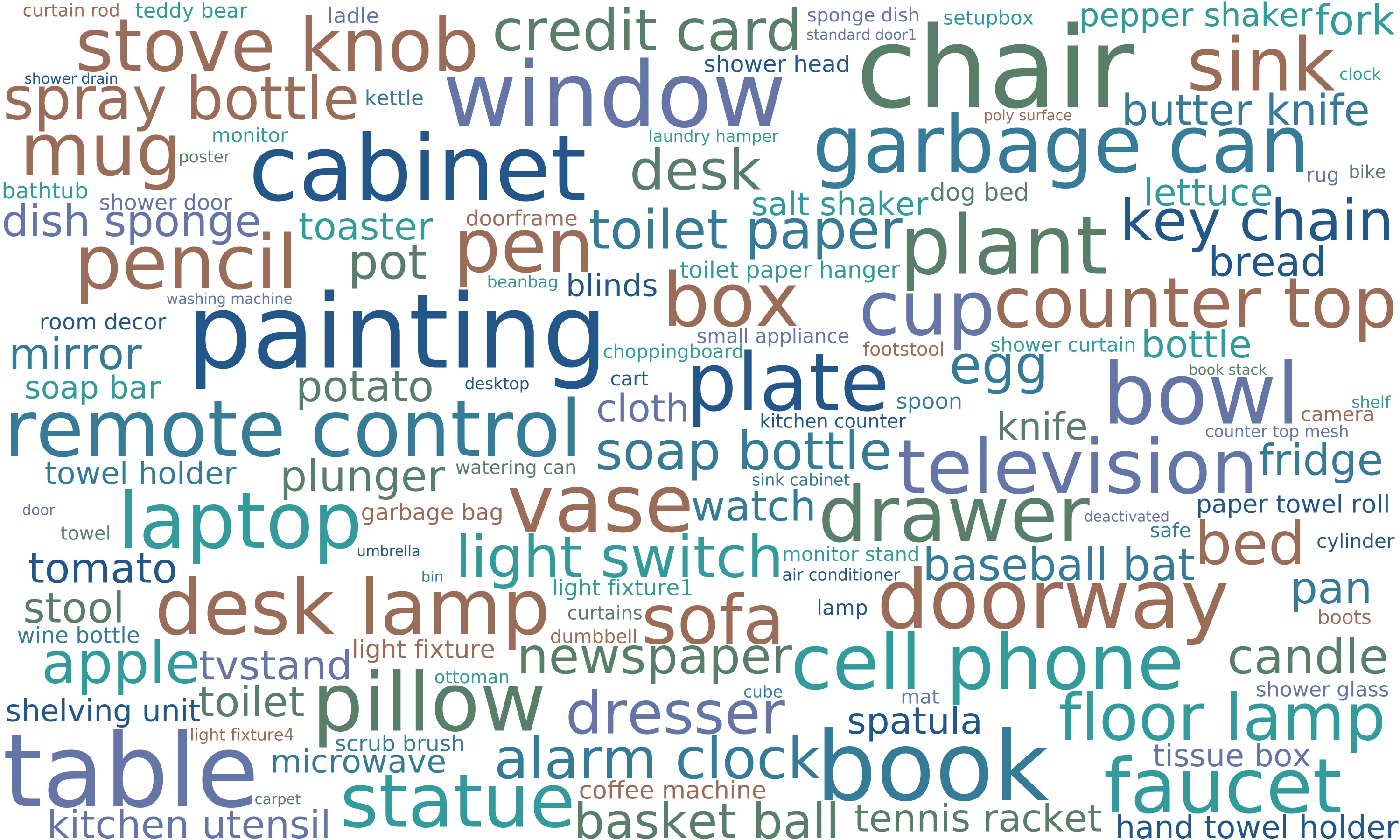}
    \caption{Wordcloud of the diverse objects in SpatialScout.}
    \label{fig:placeholder}
\end{figure}
\subsection{Benchmark Construction}
\label{sec:method_construction}

\paragraph{Scene diversity.}
Explicit geometric annotations and reproducible configurations support object-level scene evaluation~\cite{ir3d,sceneact}.
We assemble 80 indoor scenes rendered with SAPIEN~\cite{sapien} through ManiSkill3~\cite{maniskill3}: 32 iTHOR~\cite{ai2thor}, 12 RoboTHOR~\cite{robothor}, 10 ArchitecTHOR~\cite{procthor}, 6 ReplicaCAD~\cite{habitat2}, and 30 ProcTHOR~\cite{procthor} scenes.
The catalog spans individual rooms and apartments, retaining the source layouts and target placements; each source scene appears once.

\paragraph{Complexity.}
Each scene contributes 30--40 targets.
The catalog contains 12 scenes with 35 targets, and 68 with 40, totaling 3,410 instances.
The target-region radius ranges from 1.95 to 9.23\,m.
Across the catalog, 1,489 targets (43.3\%) belong to equivalence groups with multiple instances in the same scene.
Target count controls inventory scale, while spatial extent and repeated categories vary the localization and association demands.
\begin{figure*}[!t]
    \centering
    \includegraphics[width=1\linewidth]{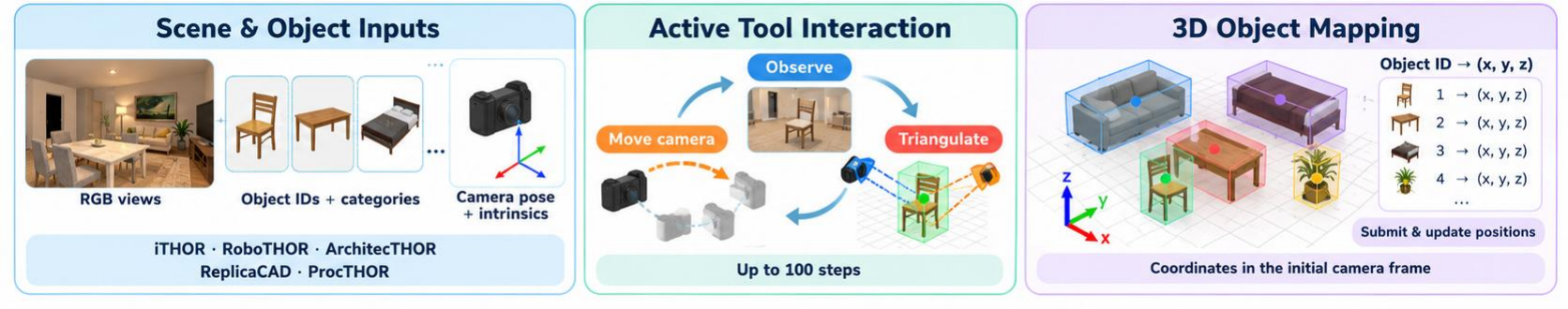}
    \caption{Evaluation of VLM agents on SpatialScout. A VLM agent navigates through the 3D space by controlling a camera, and applies observation and mapping tools to finally submit the positions of identified objects.}
    \label{fig:main}
\end{figure*}
\paragraph{Selection and initialization.}
Preparation prioritizes candidates with known semantic categories, then orders them by distance to the initial camera, using identifier order to break ties.
This rule aligns the inventory with recognizable categories and concentrates targets around the exploration origin.
Candidates must have rendered geometry, and each scene must supply the requested target count.
Candidates outside the selected target subset are hidden in rendered observations; selected targets retain their source placements.
Loaded object counts are checked against the source configuration with scene-builder exclusions accounted for.

Each catalog entry fixes the source configuration, initial pose, and target count.
Preparation saves the inventory, geometric annotations, and initial view, and verifies agreement between the camera-pose axes and projection matrix.  Appendix~\ref{sec:protocol_details} specifies the manifest and cache checks.

\section{Evaluation}
\label{sec:evaluation}
\label{sec:experiments}

\subsection{Evaluation Protocol}
\label{sec:method_protocol}
\label{sec:method_interaction}
\label{sec:experiments_setup}
\paragraph{Setup}
We evaluate seven models on a common set of 80 catalog scenes, with one episode per model--scene pair: GPT-6 Astra~\cite{openai2026gpt6astra}, GPT-6 Sol~\cite{openai2026gpt6sol}, Claude Opus~5.5~\cite{claude2026opus55}, Claude Sonnet 5.5~\cite{claude2026sonnet55}, Kimi-K3~\cite{team2026kimi}, Qwen3.8-Max~\cite{qwen2026-38max}, and DeepSeek-V4.1-Flash~\cite{ds2026-41flash}. We impose no explicit limit on output or reasoning tokens.

\paragraph{Interaction}
All agents use the ten tools in Table~\ref{tab:method_tools} through a shared interface~\cite{embodiedbench}.
Starting from one RGB image, the agent queries the inventory and calibration, acquires further views, and submits or revises object centers (see Fig.~\ref{fig:main}).
Each accepted submission replaces the previous estimate for that identifier.
The validator checks identifiers, finite coordinates, and the public bound $\|\widehat{\mathbf{p}}_i\|_2\leq r+0.5$\,m.
All agents receive the same task prompt and tool descriptions.
The prompt requests visually supported estimates and, for multi-view episodes, encourages submission after the first two views, followed by additions and revisions during exploration.
Submissions return the current accepted map, which the agent can also inspect with an empty batch.

\paragraph{Camera movement}
Relative translations follow the current horizontal forward/right directions and world vertical; positive yaw is counterclockwise viewed from above, and positive pitch looks upward.
New RGB observations are acquired through explicit captures; movement bounds are specified in Appendix~\ref{sec:protocol_details}.

\paragraph{Image coordinates}
Explicit captures render $512\times512$ RGB images, delivered as JPEGs with longest side at most 384 pixels.
Correspondences are expressed on the original image grid.
Let $(W, H)$ and $(W_d, H_d)$ denote the original and delivered dimensions.
Both grids use zero-indexed pixel centers, increasing rightward and downward.
Each image message specifies the conversion
\begin{equation}
    u=\left(u_d+\tfrac12\right)\frac{W}{W_d}-\tfrac12,\qquad
    v=\left(v_d+\tfrac12\right)\frac{H}{H_d}-\tfrac12.
    \label{eq:method_pixels}
\end{equation}
Triangulation pairs the converted pixel with the original intrinsic matrix $K$.
Each captured frame is stored with its identifier, pose, dimensions, and intrinsics.

\paragraph{Geometric support}
The triangulation tool estimates a 3D point from agent-marked correspondences in at least two calibrated frames~\cite{hartley2004}.
For camera origins $\mathbf{o}_k$ and unit viewing directions $\mathbf{d}_k$ in the task frame, define $A_k=I-\mathbf{d}_k\mathbf{d}_k^\top$.
The tool minimizes squared distances to the viewing lines:
\begin{equation}
    \widehat{\mathbf{p}}_{\mathrm{tri}}=\arg\min_{\mathbf{p}}\sum_k\|A_k(\mathbf{p}-\mathbf{o}_k)\|_2^2.
    \label{eq:method_triangulation}
\end{equation}
Queries failing baseline, conditioning, or forward-ray checks are rejected; Appendix~\ref{sec:protocol_geometry} gives the thresholds.
The tool reports the root-mean-square perpendicular line residual and maximum pairwise ray angle, with cautions for weak parallax, large residuals, or estimates outside the public range.
These diagnostics measure the geometric consistency of the supplied correspondences.
The agent must still determine whether the marked pixels identify the same object feature and how the estimated point relates to the object's center; a low residual alone establishes neither.
\begin{table*}[t]
    \centering
    \small
    \setlength{\tabcolsep}{5pt}
    \caption{Object-micro localization results on 80 scenes. All values are percentages, and threshold columns use strict IoU comparisons. The $>k\%$ columns measure the percentage of box overlaps that are greater than k\%.}
    \begin{tabular}{@{}llcccccccc@{}}
        \toprule
        \multirow{2}{*}{\textbf{Model}} & \multirow{2}{*}{\textbf{Effort}} & \multicolumn{4}{c}{\textbf{All targets}} & \multicolumn{4}{c}{\textbf{Submitted targets}} \\
        \cmidrule(lr){3-6}\cmidrule(l){7-10}
        & & \textbf{Mean IoU} & \textbf{$>0\%$} & \textbf{$>50\%$} & \textbf{$>90\%$}
              & \textbf{Mean IoU} & \textbf{$>0\%$} & \textbf{$>50\%$} & \textbf{$>90\%$} \\
        \midrule
        \includegraphics[height=0.9em]{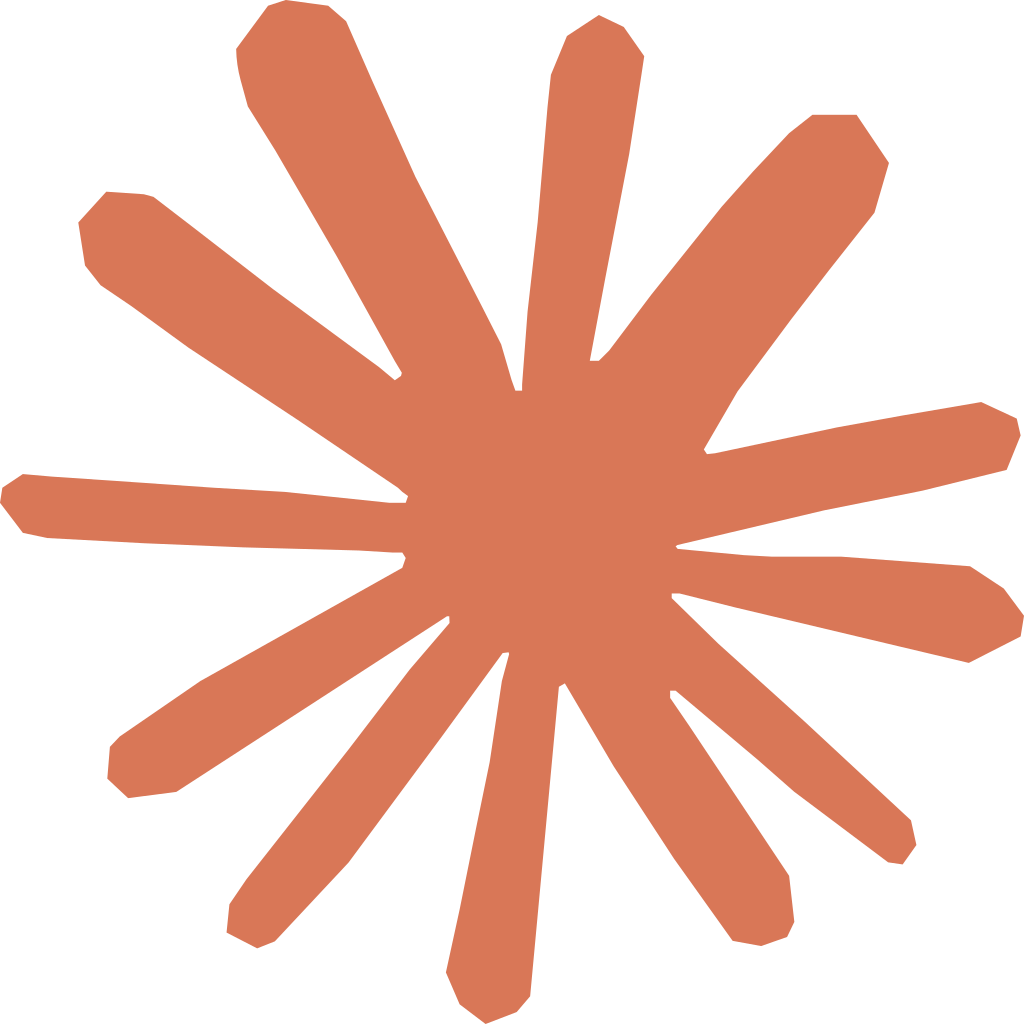} Claude Opus 5.5 & \textit{xhigh} & \textbf{41.71} & \textbf{77.36} & \textbf{47.23} & \textbf{3.25} & \textbf{46.69} & \textbf{86.60} & \textbf{52.87} & \textbf{3.64} \\
        \includegraphics[height=0.9em]{figs/claude-color.png} Claude Sonnet 5.5 & \textit{xhigh} & 32.26 & 67.13 & 33.60 & 1.40 & 38.99 & 81.14 & 40.61 & 1.69 \\
        \includegraphics[height=0.9em]{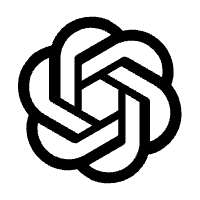} GPT-6 Astra & \textit{xhigh} & 33.16 & 67.79 & 35.15 & 1.62 & 41.16 & 83.86 & 42.69 & 2.09 \\
        \includegraphics[height=0.9em]{figs/openai-fill.png} GPT-6 Astra & \textit{medium} & 31.18 & 67.73 & 31.55 & 1.27 & 37.16 & 80.72 & 37.60 & 1.52 \\
        \includegraphics[height=0.9em]{figs/openai-fill.png} GPT-6 Sol & \textit{xhigh} & 20.83 & 51.85 & 18.95 & 0.48 & 28.65 & 71.31 & 26.06 & 0.66 \\
        
        \includegraphics[height=0.9em]{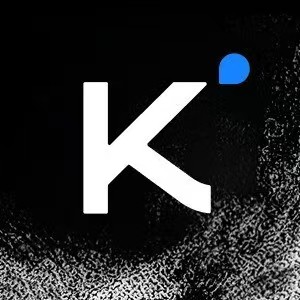} Kimi-K3 & xhigh & 4.22 & 19.36 & 2.77 & 0.00 & 6.55 & 30.03 & 4.29 & 0.00 \\
        \includegraphics[height=0.9em]{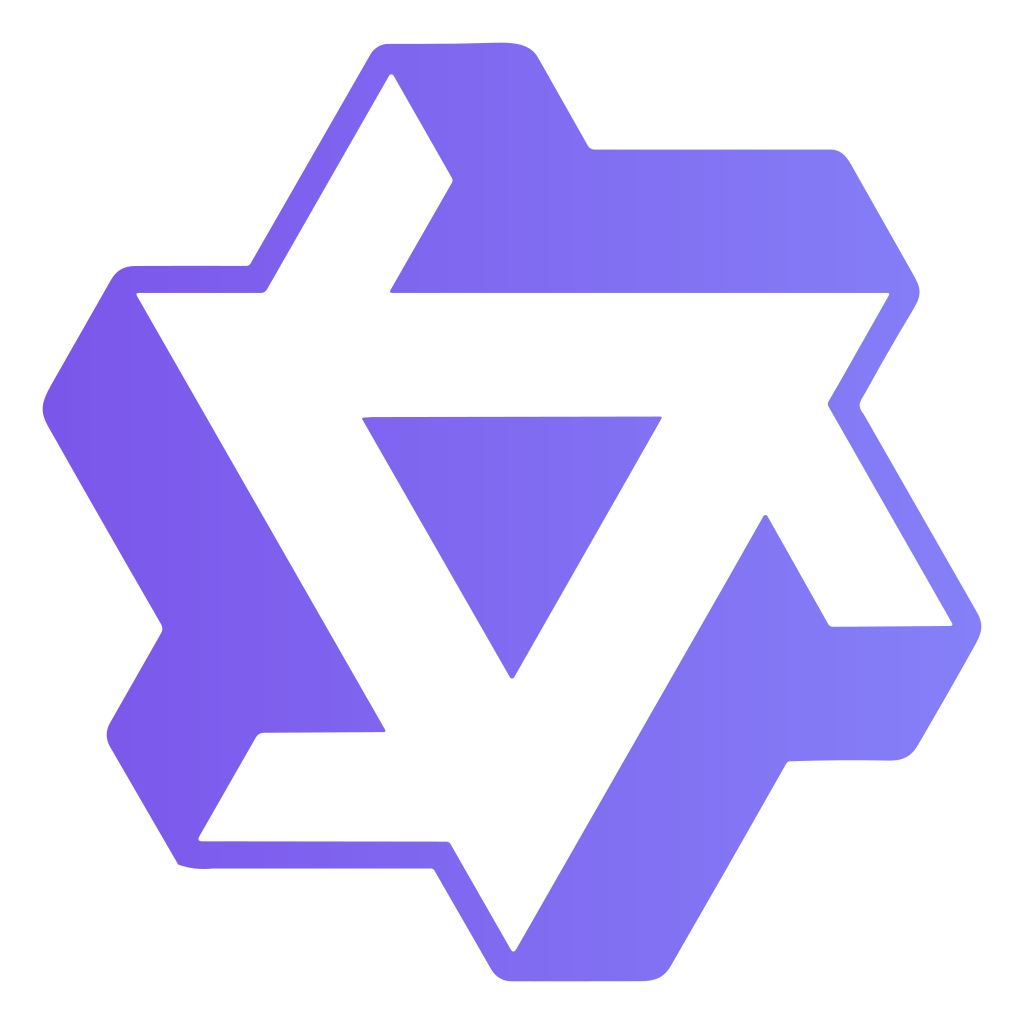} Qwen3.8 Max & \textit{xhigh} & 6.80 & 24.36 & 4.52 & 0.00 & 12.14 & 43.49 & 8.07 & 0.00 \\
        \includegraphics[height=0.9em]{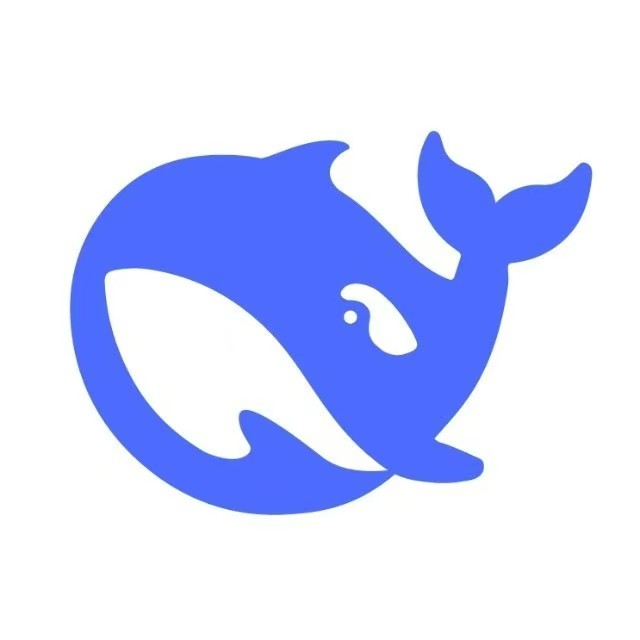} DeepSeek V4.1 Flash & \textit{xhigh} & 3.48 & 19.04 & 0.92 & 0.00 & 5.80 & 31.74 & 1.54 & 0.00 \\
        \bottomrule
    \end{tabular}
    
    \label{tab:main_results}
\end{table*}

\paragraph{Budgets and stopping.}
Each episode allows $B=100$ model calls and $V=100$ acquired images, including the initial image.
A model call may request multiple tools, executed sequentially without additional model calls; captures from the batch can inform the next call.
Reaching the capture cap leaves the remaining model calls available for reasoning and submission.
The final accepted map is scored when the agent calls \texttt{finish\_mapping}, responds without tool calls, or exhausts $B$.
Calling \texttt{finish\_mapping} freezes the map and skips subsequent tools in the same response.
Service and runtime failures are logged separately and excluded from aggregate scores.

\paragraph{Attempt records.}
Each attempt records its model and reasoning configuration, prompt, tool schema, and budgets.
Appendix~\ref{sec:protocol_details} details request accounting, rerun storage, and completion and failure reporting.

\paragraph{Map traces and replay.}
Each successful submission call records the full accepted map and model-call index.
The trace also retains capture events, frame calibration, tool arguments and results, and termination records.
An offline replay can score these snapshots to recover changes in center error and coverage as model calls and acquired images accumulate.
Reference-based scores are computed after the episode and are never returned to the agent.

\subsection{Evaluation Metrics}
\label{sec:method_evaluation}
\label{sec:experiments_metrics}

We report mean center error (meters) after minimum-distance one-to-one matching within equivalence groups~\cite{crouse2016}, together with inventory coverage.
For scale-sensitive localization, we compute 3D IoU by translating each candidate target's reference box to the predicted center.
Mean IoU uses the distance assignment; hit rates use a separate within-group hit-maximizing assignment at each strict threshold $\tau\in\{0,0.5,0.9\}$.
Overlap scores are pooled across scenes and reported over all targets (omissions count as zero) and submitted targets.
Appendix~\ref{sec:protocol_scoring} gives the scoring details.


\begin{figure*}
    \centering
    \includegraphics[width=1\linewidth]{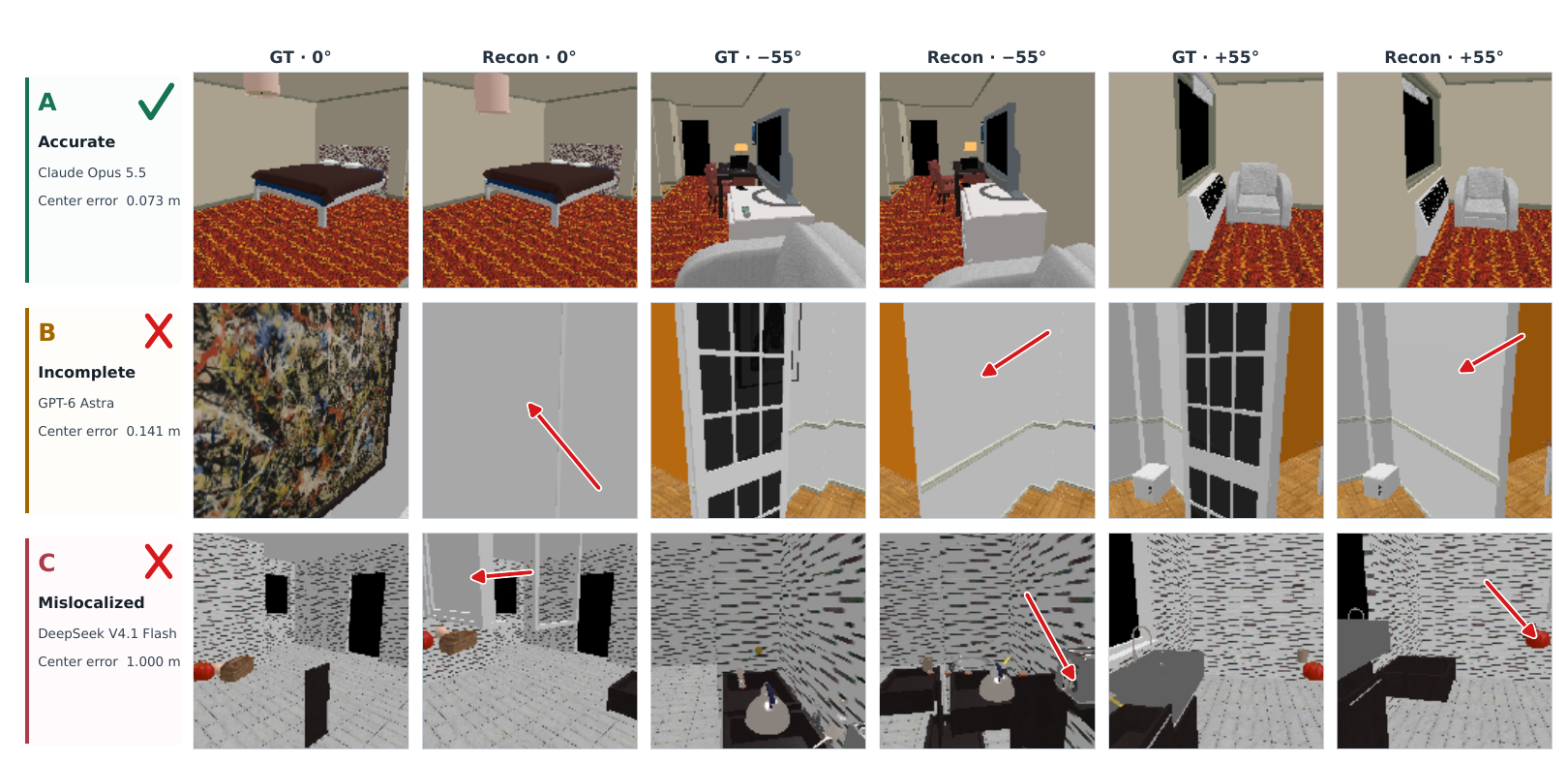}
    \caption{Examples of accurate, incomplete, and mislocalized object-center maps. The red arrows indicate a missing or mislocalized object in the reconstructed scene.}
    \label{fig:example}
\end{figure*}
\section{Results}
\subsection{Quantitative Results}
\label{sec:evaluation_results}

\paragraph{Localization and completeness.}
Table~\ref{tab:main_results} reports object-micro overlap scores for the common 80-scene comparison.
Claude Opus~5.5 achieves an all-target mean IoU of 41.71\%, compared with 33.16\% for GPT-6 Astra; their submitted-object means are 46.69\% and 41.76\%, respectively.
Claude Opus~5.5 has higher mean overlap both over the full inventory and among accepted estimates.
The decrease from submitted-object to all-target mean IoU shows how omissions reduce each model's overall mapping score.

\paragraph{Localization precision.}
Positive-overlap recall is 77.36\% for Claude Opus~5.5 and 67.79\% for GPT-6 Astra.
At IoU $>0.5$, recall is 47.23\% and 35.15\%, respectively; at IoU $>0.9$, it is 3.25\% and 1.62\%.
These differences distinguish coarse placement from precise center alignment relative to object extent.
High-IoU matches remain uncommon even among submitted objects, with hit rates at IoU $>0.9$ of 3.64\% for Claude Opus~5.5 and 2.09\% for GPT-6 Astra.

\subsection{Qualitative Analysis}
\label{sec:evaluation_qualitative}

Figure~\ref{fig:example} contrasts accurate, incomplete, and mislocalized maps.
For each comparison, the evaluator places position-matched reference meshes at predicted centers, preserves their shapes and orientations, and hides omitted targets.
Views at the initial yaw and offsets of $-55^{\circ}$ and $+55^{\circ}$ display the resulting placement and completeness from three directions.
\begin{figure*}
    \centering
    \includegraphics[width=1\linewidth]{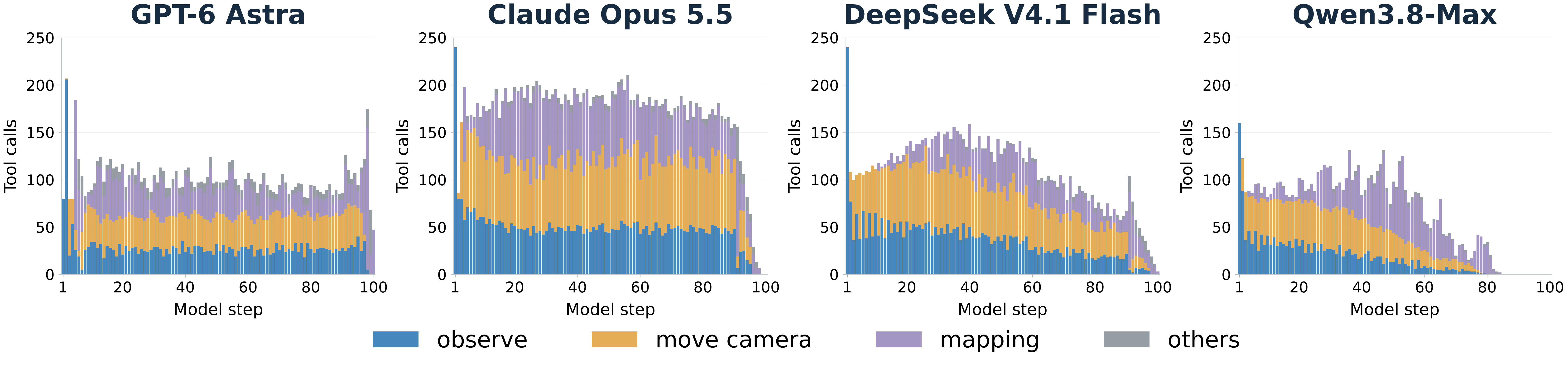}
    \caption{Tool-use patterns across interaction steps for four models. Claude Opus 5.5 makes the most tool calls, whereas GPT-6 Astra achieves competitive mapping performance with fewer calls.}
    \label{fig:tool_freq_turn}
\end{figure*}
In example~A, Claude Opus~5.5 submits all 35 targets with a center error of 0.073\,m and an all-target mean IoU of 0.530.
The reconstructed views closely align with the reference arrangement.
Example~B illustrates omissions: the GPT-6 Astra statistics cover two scenes with 65 of 80 targets submitted and a center error of 0.141\,m among submitted objects.
The comparison views make these omissions explicit.
Example~C illustrates mislocalization: DeepSeek V4.1 Flash submits all 40 targets, but the center error is 1.000\,m and all-target mean IoU is 0.034.
Together, these cases show why complete submission and accurate placement must be evaluated jointly.

\begin{figure}
    \centering
    \includegraphics[width=1\linewidth]{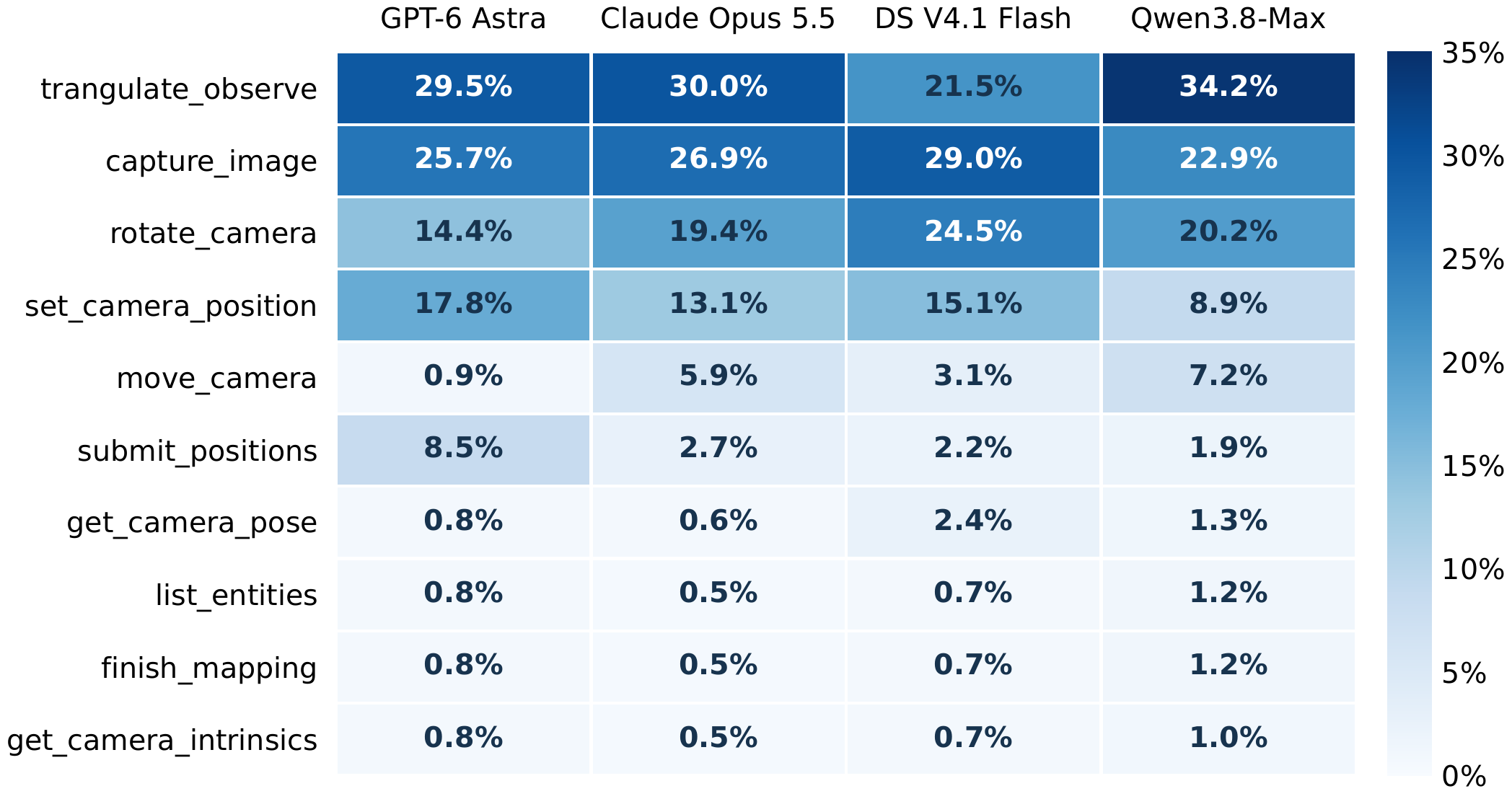}
    \caption{Illustration of tool-call frequency of multiple models.}
    \label{fig:tool_freq}
\end{figure}

\begin{table}[t]
\centering
\small
\caption{Success factors and failure patterns across 40 trajectories.
 Each trajectory may exhibit multiple factors (non-exclusive).}
\label{tab:trajectory_causes}
\begin{tabular}{@{}lp{0.62\columnwidth}r@{}}
\toprule
\textbf{Outcome} & \textbf{Factor} & \textbf{Total} \\
\midrule
Success & \textbf{S}ufficient multi-view grounding & 6 \\
        & \textbf{I}ncremental submission with continued coverage expansion & 6 \\
        & \textbf{C}ontinued refinement & 6 \\
\midrule
Failure & \textbf{I}ncomplete object coverage & 26 \\
        & \textbf{S}uspected correspondence errors despite low triangulation residuals & 21 \\
        & \textbf{U}nresolved inconsistencies in geometric measurements & 13 \\
        & \textbf{E}arly termination before coverage & 13 \\
        & \textbf{P}osition estimates lacking triangulation support & 6 \\
        & \textbf{E}xtensive observation with limited localization progress & 5 \\
        & \textbf{L}ate-stage refinement without recovering missing objects & 3 \\
\bottomrule
\end{tabular}
\end{table}
\section{Discussions}


\paragraph{Tool-calling Analysis}
We present the models' tool-calling behaviors in Fig.~\ref{fig:tool_freq_turn} and Fig.~\ref{fig:tool_freq}. These visualizations reveal differences in behaviors that provide insight into their ability to understand 3D space.

First, Claude-Opus-5.5 makes the largest number of tool calls. We cap each trajectory at 100 tool-calling turns, and a single turn may include multiple tool calls. Its extensive tool use appears to benefit its performance on SpatialScout by allowing it to gather and integrate more information about the scene.

Second, GPT-6 Astra and Claude-Opus-5.5 maintain relatively balanced use of observation, camera movement, and mapping. By contrast, DS-V4.1-Flash and Qwen3.8-Max tend to observe less as the interaction progresses. This pattern may indicate difficulty maintaining and updating a reliable spatial representation: without observing consistently, a model must rely on earlier views, which may not contain sufficient information about newly encountered objects.

Third, DeepSeek-V4.1-Flash and Qwen3.8-Max spend many turns moving the camera but relatively few observing and mapping. This suggests that they have greater difficulty selecting informative viewpoints and converting visual observations into a coherent map. Such behavior adds interaction overhead, while uninformative views also consume context because images require substantial context length.
Taken together, these behavioral differences suggest that SpatialScout distinguishes models not only by their final localization accuracy, but also by how effectively they perceive, explore, and represent the 3D scene.

\paragraph{Failure Analysis}
We analyze 40 trajectories from GPT-6 Astra, Claude Opus 5.5, DeepSeek V4.1 Flash, and Qwen3.8-Max on the same ten selected tasks, all using xhigh reasoning effort. We present the results in Table~\ref{tab:trajectory_causes}. For this case study, reconstruction success requires at least 90\% object coverage and a mean center error of at most 0.2 m over matched submitted objects. Six trajectories meet both criteria. All six exhibit extensive multi-view geometric grounding, incremental submission that expands coverage, and subsequent refinement of submitted positions. Among the 34 unsuccessful trajectories, incomplete coverage is the most common failure pattern, appearing in 26 cases. In 21 cases, at least one submitted position remains substantially inaccurate despite a low triangulation residual, suggesting incorrect visual correspondences or object-center selection that geometric consistency alone cannot resolve. Thirteen cases retain inaccurate positions associated with geometric inconsistency warnings, while another overlapping set of 13 cases terminates with insufficient coverage despite having at least ten response rounds remaining. Six cases submit a substantial fraction of positions without recorded tool-based triangulation, five involve extensive observation but limited localization progress, and three end with revisions to existing estimates without recovering missing objects. These patterns suggest that effective mapping depends on converting observations into reliable object correspondences, and checking uncertain estimates before the final termination.

\begin{table}[t]
\centering
\small
\caption{Geometric grounding on 80 shared tasks. Both models use the xhigh setting. Triangulation support is measured over submitted objects, warnings over successful triangulation calls, and median center error over submitted objects.}
\label{tab:astra_deepseek_grounding}
\begin{tabular}{p{0.59\linewidth}rr}
\toprule
\textbf{Metric} & Astra & DS-V4.1 \\
\midrule
\textbf{S}ubmitted objects with prior successful triangulation & \textbf{99.33\%} & 63.48\% \\
\textbf{T}riangulation with geometric warnings & \textbf{2.69\%} & 38.65\% \\
\textbf{M}ean image captures per task & 32.41 & 39.63 \\
\textbf{M}edian object center error (cm) & \textbf{7.08} & 68.17 \\
\bottomrule
\end{tabular}
\end{table}

\begin{figure}
    \centering
    \includegraphics[width=1\linewidth]{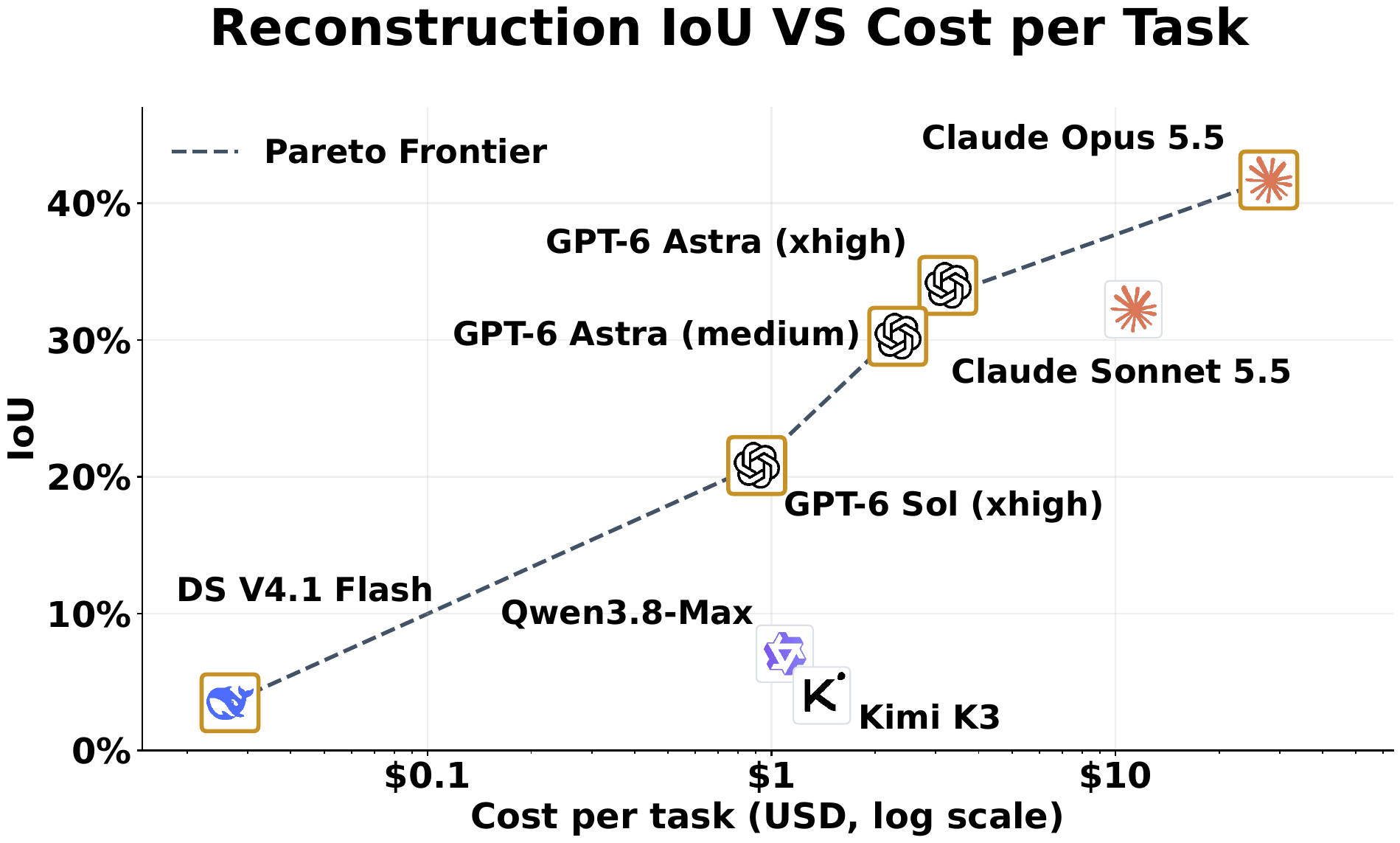}
    \caption{All-target mean IoU vs. average inference cost per task.}
    \label{fig:cost}
\end{figure}

\paragraph{Trajectory Analysis}

We use Codex powered by GPT-6 Astra to analyze both models' trajectories across 80 shared tasks, as summarized in Table~\ref{tab:astra_deepseek_grounding}. We select GPT-6 Astra and DeepSeek-V4.1-flash because they make similar numbers of tool calls but achieve substantially different localization performance. GPT-6 Astra captures fewer images yet provides triangulation support for more submitted objects, with fewer geometric warnings. The triangulation tool requires fine-grained spatial perception to identify corresponding image points across views; DeepSeek-V4.1-Flash's inconsistent measurements suggest difficulty meeting this requirement, despite capturing more images. These findings suggest that more effective geometric tool use contributes to GPT-6 Astra's stronger localization performance.

\paragraph{Cost Analysis}
Figure~\ref{fig:cost} plots the mean IoU across all targets against the average inference cost (in USD, on a logarithmic scale), together with the Pareto frontier. Claude Opus 5.5 achieves the highest IoU, but also incurs the highest cost (28.85 USD per task), consistent with findings from other long-horizon agentic benchmarks~\cite{swebench,repozero}. Overall, the evaluated open-source models offer limited advantages in the accuracy--cost trade-off. Among them, only DeepSeek-V4.1-Flash costs less than GPT-6 Sol; Qwen3.8-Max and Kimi-K3 are both more expensive and achieve lower scores on all IoU metrics (Table~\ref{tab:main_results}).

\section{Related Work}
\label{sec:related_work}

\paragraph{Agentic inverse graphics and scene construction.}
IR3D-Bench evaluates scene understanding by reconstructing a scene from a supplied image~\cite{ir3d}.
BlenderGym studies code-based graphics editing, including how generation, verification, and their compute allocation affect performance~\cite{blendergym}.
BlenderAlchemy combines edit generation with visual evaluation~\cite{alchemy}, while VIGA and BlenderBench examine interleaved execution, inspection, memory, and exploratory editing~\cite{viga}.
SceneActBench evaluates executable multi-object outputs across several tasks; its reconstruction task builds furniture meshes in an initially empty scene from supplied calibrated views~\cite{sceneact}.
These works establish construction and editing as meaningful tests of visual understanding.
Our setting emphasizes acquisition of new evidence from a hidden target environment and inspection and revision of an explicit object-center map.

\paragraph{Active reconstruction and next-best-view planning.}
Selecting informative observations has a substantial history in three-dimensional perception.
SCONE predicts surface coverage gain~\cite{scone}, and MACARONS couples RGB-based mapping with online coverage anticipation~\cite{macarons}.
GenNBV learns a generalizable view-selection policy~\cite{gennbv}.
NARUTO, ActiveSplat, and ActiveGAMER integrate exploration with neural scene mapping~\cite{naruto,activesplat,activegamer}; FisherRF uses information-based view selection for radiance fields~\cite{fisherrf}.
More directly, AIR-Embodied uses multimodal reasoning to plan reconstruction actions~\cite{air}.
AREA3D combines geometric perception and vision-language guidance, and introduces an evaluation covering object- and scene-level active reconstruction~\cite{area3d}.
SightCAD connects feasible view selection to executable CAD generation for scene-embedded objects~\cite{sightcad}.
These systems provide important planning and reconstruction references.
Our benchmark compares general-purpose agents under shared tools and observation and model-call budgets, recording how their evidence acquisition and map updates interact.

\paragraph{Spatial and embodied benchmarks.}
VSI-Bench and SITE assess spatial reasoning from supplied visual observations~\cite{vsi,site}.
OpenEQA includes both episodic-memory and active-exploration settings~\cite{openeqa}, and Explore until Confident studies exploration and stopping for embodied question answering~\cite{exploreeqa}.
E3VS-Bench evaluates viewpoint-dependent evidence acquisition through camera control~\cite{e3vs}; ESI-Bench covers spatial tasks that require active observation and interaction~\cite{esi}.
S$^3$-Bench combines streaming spatial questions with memory and additional exploration~\cite{s3}.
EmbodiedBench evaluates visual agents across multiple environments and action levels~\cite{embodiedbench}.
Our evaluation complements these tasks by scoring the explicit object-center map maintained over an episode. Compared with QA pairs, our tasks require VLM agents to output the precise 3D coordinates of the objects.

\paragraph{Memory, geometry tools, and process diagnosis.}
ConceptFusion and ConceptGraphs build semantically rich three-dimensional representations through multimodal fusion and cross-view association~\cite{conceptfusion,conceptgraphs}.
3D-Mem uses scene memory for embodied exploration and reasoning~\cite{mem3d}, while SpaMEM separates perceptual grounding from temporal state maintenance~\cite{spamem}.
DUSt3R and VGGT provide geometric inference tools~\cite{dust3r,vggt}, and NeRF and Gaussian splatting offer scene representations for rendering and reconstruction~\cite{nerf,gaussians}.
Alongside BlenderGym, 3DCodeBench investigates inference-time computation in graphics tasks~\cite{blendergym,code3d}.
Building on these perspectives, we control image acquisition and model calls with separate budgets.
Shared tools, inventory-only reference policies, and replay of accepted-map updates support comparisons of final localization.

\paragraph{Long-horizon agentic benchmarks.}
Long-horizon agentic tasks require sustained reasoning and repeated tool use, often over multiple interaction rounds, and remain a major challenge for deploying LLM-based systems in real-world settings. AgentBench and GAIA evaluate interactive decision-making and general-purpose tool use~\cite{agentbench,gaia}. WebShop and WebArena focus on web interaction, while VisualWebArena and OSWorld extend evaluation to visually grounded web and desktop environments~\cite{webshop,webarena,visualwebarena,osworld}. The $\tau$-bench and DySQL-Bench examine tool use in settings with evolving user interactions~\cite{taubench,dysql}. SWE-bench, Terminal-Bench, NL2Repo, and RepoZero assess repository repair, command-line problem solving, and repository generation using executable checks~\cite{swebench,terminalbench,ding2025nl2repo,repozero}. AgenticRAGTracer complements these benchmarks with hop-level diagnosis of agentic retrieval processes~\cite{agenticragtracer}. SpatialScout is one of the benchmarks evaluating the spatial understanding and long-horizon capability of VLM agents.
\section{Conclusion}
We introduced SpatialScout, a benchmark for evaluating the spatial intelligence of VLM agents through active object-level 3D mapping. By requiring agents to autonomously select viewpoints, integrate visual evidence, and estimate object positions in a shared coordinate frame, SpatialScout evaluates the complete process of constructing metric spatial representations. Our experiments reveal that frontier VLMs can recover partial scene layouts but still struggle to achieve both comprehensive coverage and precise localization. Even the strongest evaluated model attains a mean IoU of 41.71\%, with only 3.25\% of targets exceeding an IoU of 0.9. These findings highlight the difficulty of translating visual observations into reliable geometric estimates through sustained interaction. SpatialScout provides a testbed for advancing active perception, cross-view reasoning, and persistent spatial memory, supporting the development of agents capable of building accurate and complete representations of 3D environments.
{
    \small
    \bibliographystyle{ieeenat_fullname}
    \bibliography{main}
}


\clearpage
\appendix
\begin{figure}
    \centering
    \includegraphics[width=0.8\linewidth]{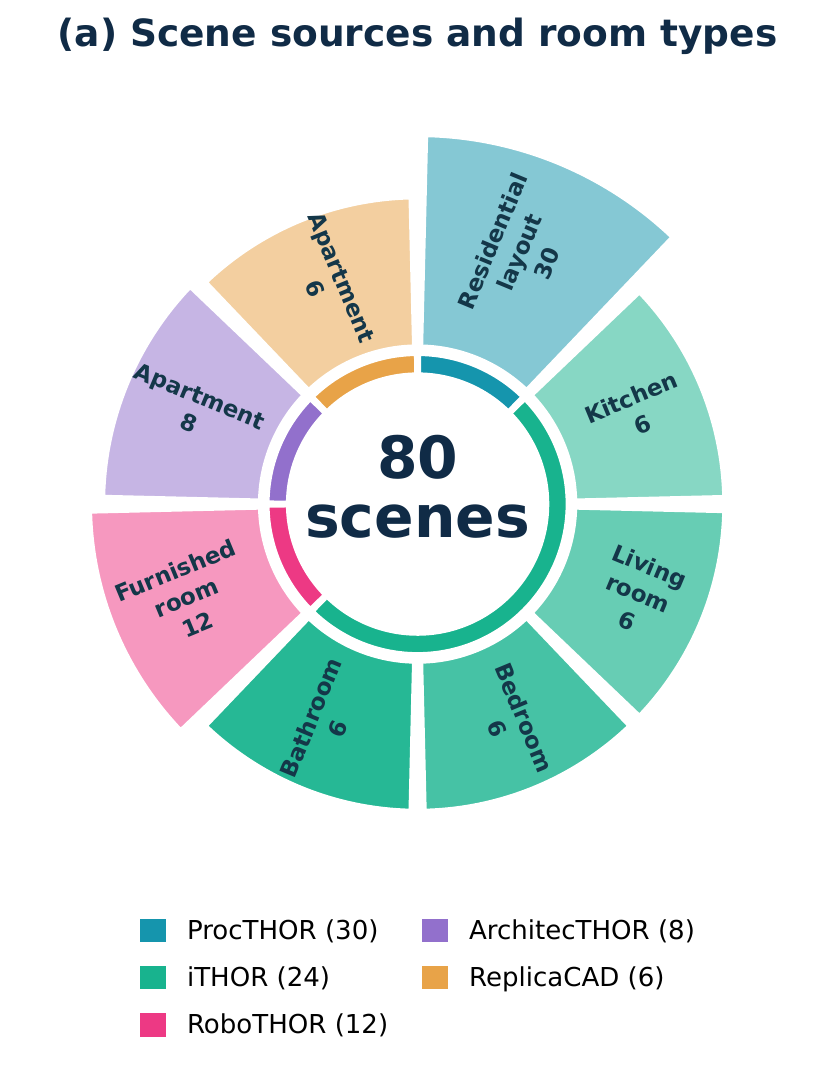}
    \caption{Visualization of scene distribution in SpatialScout}
    \label{fig:distribution}
\end{figure}

\section{Protocol}
\label{sec:protocol_details}

\begin{figure}
    \centering
    \includegraphics[width=0.75\linewidth]{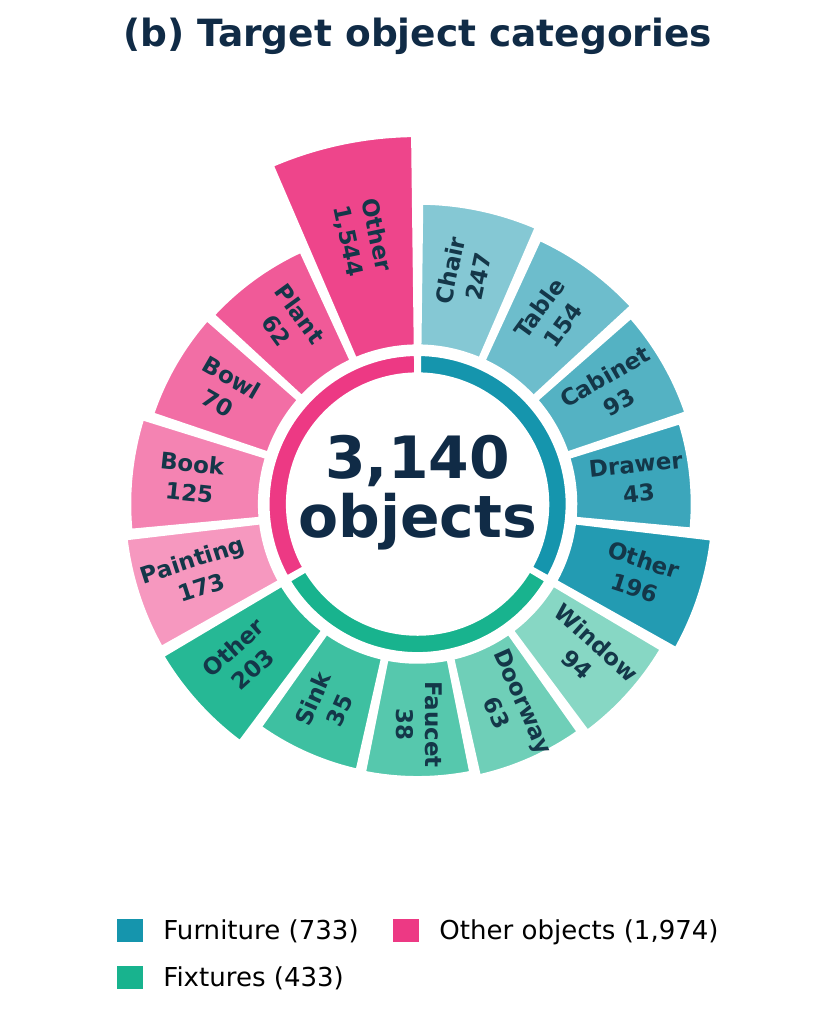}
    \caption{Visualization of object distribution in SpatialScout}
    \label{fig:distribution_object}
\end{figure}

\paragraph{Scene preparation.}
Catalog entries identify the scene builder, source configuration, initial world position and yaw--pitch, and target count.
The preparation script verifies the loaded source identity and applies the target-selection rule in Sec.~\ref{sec:method_construction}.
The default configuration uses seed 0, $512\times512$ rendering, CPU simulation and rendering backends, and the minimal shader setting.
The delivered image uses JPEG quality 82 and the pixel mapping in Eq.~\eqref{eq:method_pixels}.
The preparation manifest records the random seed, render size, simulation and rendering backends, input-image encoding, dependency versions, and SHA-256 fingerprints of the source configuration, benchmark and simulator code, and asset tree.
Cached preparations are reused only when their fingerprints, initial camera poses, and target counts agree with the requested configuration.
Each run records the preparation fingerprint, linking the attempt to its scene and data configuration.
The public entity inventory, evaluator annotations, and initial view are saved as separate artifacts.
Each frame also retains its original-image filename. The overall data distribution is presented in Fig.~\ref{fig:distribution} and Fig.~\ref{fig:distribution_object}.

\paragraph{Movement.}
Absolute positions are transformed from the task frame to world coordinates by $\mathbf{o}_0+R_0\mathbf{p}$, with each world coordinate bounded by $\pm15$\,m.
Relative translations have length at most 1\,m; yaw and pitch increments have magnitude at most $180^{\circ}$ per axis, final pitch is clamped to $[-85^{\circ},85^{\circ}]$, and roll remains fixed.
Section~\ref{sec:method_protocol} specifies the movement axes and explicit capture interface.

\paragraph{Prompts and budgets.}
Chat Completions and Responses use a shared episode loop, task prompt, and tool descriptions.
Section~\ref{sec:method_protocol} describes the common submission guidance and map-inspection interface.
The prompt also specifies the actual model-call and capture budgets, the fixed task axes, and the world-AABB center convention.
When $V=1$, it requests estimates from the initial image and its calibration.
Budget reminders are added before model requests with 10 and 3 calls remaining.
Actions within a single response are specified before the corresponding visual feedback is seen.
Rejected capture requests do not consume additional images. The default user prompt and the tool descriptions are presented in Fig.~\ref{fig:task_prompt} and Fig.~\ref{fig:tool_instructions}.

\paragraph{Attempt accounting.}
The standard batch uses a 180-second timeout per HTTP request, with no explicit temperature setting.
Each episode attempt records the model identifier, API mode, reasoning configuration, full task prompt, tool schema, and actual budgets.
Retriable HTTP or network failures receive up to three request attempts with 1- and 2-second backoffs.
Logical model calls and HTTP request attempts have separate counters.
The batch runner stores reruns in distinct attempt directories, preserving previous maps and traces.
Completed episodes are labeled by explicit finish, a response without tool calls, or turn-budget exhaustion.
Service and runtime failures retain available maps and diagnostic records under the accounting policy in Sec.~\ref{sec:method_protocol}.
Batch reports retain completed-episode counts and failure counts and rates alongside aggregate scores.
Comparisons use a specified scene list and report the completed count for each model.
\section{Reproducibility Details}
\subsection{Multi-view Point Estimation}
\label{sec:protocol_geometry}
\label{sec:method_geometry}
For a marked original-image pixel $(u_k,v_k)$, the forward/left/up camera convention gives
\begin{equation}
    \mathbf{q}_k=\begin{bmatrix}
    1\\[-1pt]-(u_k-c_{x,k})/f_{x,k}\\[-1pt]-(v_k-c_{y,k})/f_{y,k}
    \end{bmatrix},\qquad
    \mathbf{d}_k=\frac{R_k\mathbf{q}_k}{\|R_k\mathbf{q}_k\|_2},
    \label{eq:method_ray}
\end{equation}
where $R_k$ maps camera axes into the fixed task frame.
Using the task-frame origins $\mathbf{o}_k$ and projections $A_k$ defined in Sec.~\ref{sec:method_protocol},
the normal equations of Eq.~\eqref{eq:method_triangulation} give
\begin{equation}
    \widehat{\mathbf{p}}_{\mathrm{tri}}=\left(\sum_k A_k\right)^{-1}\sum_k A_k\mathbf{o}_k.
\end{equation}
A valid query requires at least two observations, maximum camera separation of at least 5\,cm, condition number of $\sum_k A_k$ at most $10^6$, and $\mathbf{d}_k^\top(\widehat{\mathbf{p}}_{\mathrm{tri}}-\mathbf{o}_k)>0$ for every contributing ray.
Queries failing these checks are rejected.
The tool reports the root-mean-square perpendicular line residual and maximum pairwise ray angle.
Valid solutions carry a caution when the angle is below $3^{\circ}$, the residual exceeds 0.1\,m, or the point lies beyond $r+0.5$\,m.

\subsection{Scoring and Trace Replay}
\label{sec:protocol_scoring}
For a scene with $N>0$ targets, let $M$ be the number of distinct target IDs in the current accepted map.
The position evaluator uses groupwise minimum-total-Euclidean-distance assignment.
Mean, median, RMSE, and mean IoU use that assignment.
Before assignment, predictions are ordered lexicographically by task-frame coordinates and targets by task-frame center and world-box side lengths, with identifiers breaking ties between geometrically identical entries.
This canonical ordering makes tied assignments reproducible under permutations of same-group submission identifiers.
The target-conditioned overlap evaluator floors each reference side length at $\epsilon=0.01$\,m while preserving its center.
For each prediction--target pair it translates the candidate target's box to the prediction, giving a score determined by the predicted center, candidate center, and candidate extent.
At each threshold, one-to-one matching first maximizes strict threshold hits and then total overlap.
Thresholds are $\tau\in\{0,0.5,0.9\}$, with separate assignments for each threshold.
The scorer saves the metric version, minimum side length, matched IoU sum $Q$, hit counts $H_\tau$, and denominators $M,N$.

\paragraph{Aggregation and completeness.}
For evaluated scenes $\mathcal{D}$, object-micro means use $\sum_{s\in\mathcal{D}}Q_s/\sum_s M_s$ for submitted objects and $\sum_s Q_s/\sum_s N_s$ for all targets.
Replacing $Q_s$ with $H_{\tau,s}$ gives the corresponding hit rate and recall.
Scene-macro summaries instead average the per-scene ratios with equal weight.
Both mean IoU variants and overlap rates are zero for an empty map; submitted-object center errors are undefined and excluded from averages of those errors.
IoU and overlap-rate aggregates with zero denominators are also set to zero.
At scene level, and under object-micro aggregation with $C=\sum_s M_s/\sum_s N_s$,
\begin{equation}
    \overline{J}_{\mathrm{all}}=C\overline{J}_{\mathrm{sub}},\qquad
    \operatorname{Recall}_{\tau}=C\operatorname{HitRate}_{\tau}.
\end{equation}
For scene-macro reports these identities apply within each scene before averaging.

\paragraph{Omission penalty.}
With coverage $C=M/N$, the auxiliary full-scene localization loss is
\begin{equation}
    L_{\mathrm{scene}}=\frac{\sum_{i=1}^{M}e_i+\lambda(N-M)}{N},
\end{equation}
where $e_i$ is the matched center error in meters and $\lambda$ is a fixed penalty per omitted target, also in meters.
The penalty is configurable; we use the evaluator's default $\lambda=5$\,m, so each omitted target adds 5\,m to the summed error before normalization by $N$.
A larger $\lambda$ gives omissions more weight relative to matched localization error.
For an empty map, $L_{\mathrm{scene}}=\lambda$.
For a nonempty map, the loss decomposes as $L_{\mathrm{scene}}=C\bar e+(1-C)\lambda$, where $\bar e$ is mean center error.
Reporting coverage and center error alongside the loss makes the contribution of omissions explicit.
The main-table mean IoU and overlap hit/recall scores are independent of $\lambda$.

\paragraph{Object scale.}
Across the 80 evaluation scenes, 210 of 3,140 targets (6.7\%) have at least one world-box side shorter than 1\,cm and use the minimum-side regularization.
The median shortest side before regularization is 11.7\,cm.
For two identical regularized world-axis-aligned boxes displaced by $d$ along one axis of length $\ell$, with the other axes aligned and $|d|<\ell$,
\begin{equation}
    J=\frac{\ell-|d|}{\ell+|d|},\qquad
    J>\tau\ \Longleftrightarrow\ |d|<\ell\frac{1-\tau}{1+\tau}.
    \label{eq:scale_displacement}
\end{equation}
At IoU $>0.9$, the permitted displacement is below $\ell/19$, approximately 5.26\% of that axis length.
At the 1\,cm minimum side length, the single-axis tolerance is approximately 0.526\,mm.
Center error complements these scale-relative scores by measuring displacement in meters.

\paragraph{Inventory-only references.}
The origin policy submits $\widehat{\mathbf{p}}_i=\mathbf{0}$ for every inventory item.
The uniform-ball policy samples a unit direction from a normalized three-dimensional Gaussian vector and a radius $rU^{1/3}$, $U\sim\mathrm{Uniform}[0,1]$, using seed 0.
Both policies use the same groupwise scoring and cover the full inventory.
Across the same 80 evaluation scenes, their scene-macro full-scene losses are 3.113\,m and 4.701\,m, respectively.
These values provide RGB-free localization baselines using the public inventory and range prior.

\paragraph{Map-update replay.}
The replay in Sec.~\ref{sec:method_protocol} reconstructs the acquired-image count from the initial capture and subsequent successful capture events.
It scores the full map returned by every successful submission call, including empty batches used to inspect the map.
The same accepted maps can also be evaluated with omission penalties $\lambda\in\{1,2,5,10\}$\,m, separating the effect of the reporting trade-off from changes in predictions without further model inference.
Model text and captured-image files accompany the calibrated frames and tool records.
Resource records include request and execution times, counts of model calls, acquired images and tool calls, and provider-reported token usage when available.

\section{Limitations}
SpatialScout currently relies on data generated using a physics simulation engine. Although our results support the conclusions drawn in this work, simulated environments may not fully capture the complexity and variability of real-world scenes. To address this limitation, we plan to release a dataset collected in real-world scenes within the next month. We welcome any forms of collaboration.

\begin{figure*}[t]
\centering
\begin{tcolorbox}[
  title={Task Prompt},
  colback=gray!4,
  colframe=black!65,
  colbacktitle=black!65,
  coltitle=white,
  fonttitle=\bfseries,
  fontupper=\small,
  boxrule=0.6pt,
  arc=2mm,
  left=3mm, right=3mm, top=2mm, bottom=2mm,
  width=\textwidth
]
Explore this furnished scene and estimate the center of each entity's world-axis-aligned bounding box in the supplied library. Define the box in simulator world axes, take the midpoint of its bounds, then express that point in the task frame; do not recompute an axis-aligned box in the task frame. All submitted coordinates use a fixed initial-camera frame: origin at the camera's initial position, +x its initial optical forward, +y its initial left, +z its initial up. Camera pose tools return transforms in this same frame. Absolute \texttt{set\_camera\_position} arguments use this frame. Relative \texttt{move\_camera} uses current horizontal heading and world-vertical up. A silhouette center can shift between views and a surface point differs from the object's bounding-box center. Use the observed object extent to infer its bounding-box center. \texttt{triangulate\_observations} accepts ORIGINAL PNG \texttt{pixel\_xy}, with zero-based pixel centers, u right and v down. Each image message gives original/displayed dimensions and the exact displayed-to-original conversion. Only submit a position when you have visual evidence; do not invent locations for occluded items. The library includes equivalence groups. IDs within the same group are interchangeable for scoring, so submit one distinct valid ID per observed instance without worrying about their order. Submit estimates in batches using \texttt{submit\_positions}, then call \texttt{finish\_mapping}. \texttt{finish\_mapping} immediately freezes the map; later tools in that response are skipped. The architectural stage mesh and robot are outside the target library; some furniture and fixtures are separate target instances. This benchmark keeps a selected subset of targets visible; map only IDs returned by \texttt{list\_entities}. Use multiple images and camera poses. Triangulate the same physical feature seen from different positions. Start submitting visible objects after the first two views, then update and add estimates as you explore. You have at most 100 model calls; one call may request multiple sequential tools. You have at most 100 RGB captures including the initial image. After this capture cap, use the remaining model calls to reason from existing images, submit, and finish.
\end{tcolorbox}
\caption{The user prompt provided to the visual mapping agent for the default evaluation with $B=V=100$.}
\label{fig:task_prompt}
\end{figure*}

\begin{figure*}[t]
\centering
\begin{tcolorbox}[
  title={Tool Instructions},
  colback=gray!4,
  colframe=black!65,
  colbacktitle=black!65,
  coltitle=white,
  fonttitle=\bfseries,
  fontupper=\small,
  boxrule=0.6pt,
  arc=2mm,
  left=3mm, right=3mm, top=2mm, bottom=2mm,
  width=\textwidth
]
\par\medskip
\noindent\textbf{\texttt{get\_camera\_pose}.} Read camera position and full rotation matrix in the ORIGINAL camera frame. \\
\emph{Parameters:} No arguments.\par\smallskip

\noindent\textbf{\texttt{get\_camera\_intrinsics}.} Read the actual 3x3 camera intrinsic matrix, image size, and clipping planes. \\
\emph{Parameters:} No arguments.\par\smallskip

\noindent\textbf{\texttt{set\_camera\_position}.} Teleport the camera to absolute x,y,z coordinates in the ORIGINAL camera frame, in meters. \\
\emph{Parameters:} Required: \texttt{x}, \texttt{y}, \texttt{z} (numbers; meters).\par\smallskip

\noindent\textbf{\texttt{move\_camera}.} Move along the current yaw-horizontal forward/right axes and world-vertical up, in meters; vector length at most 1 meter. These relative axes differ from fixed task axes when the camera rotates or pitches. \\
\emph{Parameters:} Required: \texttt{forward}, \texttt{right}, \texttt{up} (numbers; meters).\par\smallskip

\noindent\textbf{\texttt{rotate\_camera}.} Rotate the camera by relative yaw and pitch in degrees. Positive yaw turns counterclockwise viewed from above; positive pitch looks upward. \\
\emph{Parameters:} Required: \texttt{yaw\_deg}, \texttt{pitch\_deg} (numbers; degrees).\par\smallskip

\noindent\textbf{\texttt{capture\_image}.} Capture the camera's current RGB image. The PNG is saved locally and the image is sent to your vision input. \\
\emph{Parameters:} No arguments.\par\smallskip

\noindent\textbf{\texttt{list\_entities}.} List every target entity ID and category in this room. No locations are revealed. \\
\emph{Parameters:} No arguments.\par\smallskip

\noindent\textbf{\texttt{triangulate\_observations}.} Triangulate the same physical point marked in two or more captured frames taken from different camera positions. Returns an absolute 3D point in the initial-camera frame and ray mismatch. A visible surface point or silhouette center need not be the object's bounding-box center; infer the object center using its extent. \\
\emph{Parameters:} Required: \texttt{id} (string) and \texttt{observations} (array of at least two observations). Each observation requires \texttt{frame} (captured filename, such as \texttt{frame\_0001.png}) and \texttt{pixel\_xy} (two numbers $[u,v]$ in original PNG coordinates, with zero-based pixel centers; convert displayed JPEG coordinates first).\par\smallskip

\noindent\textbf{\texttt{submit\_positions}.} Submit or update a batch of estimated geometric-center coordinates in the original-camera frame. Returns acceptance diagnostics and the current accepted map, without reference errors. Use exact entity IDs from the library. Call repeatedly to revise the map, or pass an empty batch to inspect it. \\
\emph{Parameters:} Required: \texttt{predictions} (array). Each entry requires \texttt{id} (string) and \texttt{xyz\_m} (three numbers).\par\smallskip

\noindent\textbf{\texttt{finish\_mapping}.} Immediately finish mapping with the current submitted positions. Later tool calls in this batch are not executed. Missing IDs count as missing predictions; scoring is performed after the episode. \\
\emph{Parameters:} No arguments.\par\smallskip
\end{tcolorbox}
\caption{Tool descriptions and summarized parameter schemas provided to the visual mapping agent.}
\label{fig:tool_instructions}
\end{figure*}

\end{document}